\documentclass[runningheads]{llncs}
\usepackage[T1]{fontenc}
\usepackage{graphicx,verbatim}
\usepackage{amsmath}
\usepackage[table]{xcolor}
\definecolor{lightgray}{gray}{0.92}
\usepackage{multirow}
\usepackage{booktabs}
\usepackage{wrapfig}
\usepackage{graphicx} % Required for \resizebox and \includegraphics
\usepackage{amsmath}  % Required for math symbols and \text
\usepackage{amssymb}  % Required for math fonts like \mathcal
\usepackage{caption}
\usepackage{hyperref}
\usepackage[table]{xcolor}
\usepackage{orcidlink}
\hypersetup{
    colorlinks,
    linkcolor={red!50!black},
    citecolor={blue!50!black},
    urlcolor={blue!50!black}
}
\usepackage{graphicx}
\usepackage{amsmath}
\usepackage{amssymb}
\usepackage{tikz}
\usetikzlibrary{positioning, arrows.meta, shapes.geometric, shadows, calc}
\usetikzlibrary{
    positioning, 
    arrows.meta,   % For nicer, modern arrowheads (-Latex)
    shapes.geometric, % For the trapezoid encoder/decoder shapes
    shadows,       % For depth/drop shadows
    calc           % For precise coordinate calculations if needed
}
\usepackage{pgfplots}
\pgfplotsset{compat=1.18}
\usepgfplotslibrary{groupplots}

\begin{document}
\title{The Learnability Gap in Medical Latent Diffusion}
\titlerunning{The Learnability Gap}

\author{%
Mischa Dombrowski\inst{1}\orcidlink{0000-0003-1061-8990} \and
Felix N\"utzel\inst{1}\orcidlink{0009-0000-8277-172X} \and
Bernhard Kainz\inst{1,2}\orcidlink{0000-0002-7813-5023}%
}

\index{Dombrowski, Mischa}
\index{N\"utzel, Felix}
\index{Kainz, Bernhard}

\authorrunning{M. Dombrowski et al.}

\institute{%
FAU Erlangen-N\"urnberg, Erlangen, DE, 
\email{mischa.dombrowski@fau.de}
\and
Department of Computing, Imperial College London, London, UK
}

\maketitle

\begin{abstract}
Generative data augmentation with latent diffusion models is a promising strategy for addressing class imbalance in medical imaging, yet current approaches focus on perceptual fidelity and domain-specific autoencoder fine-tuning while neglecting a more fundamental bottleneck. We identify and formalize the \emph{learnability gap}: large-scale pretrained autoencoders faithfully encode discriminative features for medical classification, as evidenced by near-lossless performance in reconstruction space, yet their latent representations are structured in ways that are difficult for classifiers to learn from. Across five autoencoder families and four medical benchmarks spanning chest radiography, dermatoscopy, computed tomography, and echocardiography, we show that this gap persists regardless of architecture, initialization strategy, or hyperparameter tuning, and that medical-domain fine-tuning of the autoencoder does not close it. To probe and partially narrow the gap, we develop noise-conditioned latent classifiers with FiLM layers and image-space distillation that offer $64{\times}$ throughput and $120{\times}$ memory gains over image-space models while serving as diagnostic tools for latent space quality. Our analysis provides a new framework for evaluating autoencoder latent spaces and identifies their structure, rather than their fidelity or domain specificity, as the primary obstacle to closing the performance gap between real and synthetic medical training data.

\keywords{Image generation \and Latent diffusion \and Representation learning \and  Long-tail classification.}

\end{abstract}

\section{Introduction}
\label{sec:intro}

Clinical datasets follow long-tailed distributions almost universally: common diagnoses dominate while rare but critical conditions populate the tail~\cite{holste_long-tailed_2022,holste_towards_2024,lin_cxr-lt_2025}. In chest radiography, ``No Finding'' can exceed 60\% of studies while pneumomediastinum appears in fewer than 0.1\%~\cite{johnson_mimic-cxr_2019}. Similar imbalances arise in dermatoscopy~\cite{codella2018skin,tschandl2018ham10000}, computed tomography~\cite{hamamci_developing_2025}, and congenital heart disease screening~\cite{vega2025cardiumcongenitalanomalyrecognition}. Models trained on such distributions systematically underdiagnose rare conditions~\cite{cao_learning_2019,kang_decoupling_2020}, and acquiring additional labeled examples is expensive and privacy-constrained~\cite{rieke2020future}.

\noindent Latent diffusion models~\cite{rombach2022high,peebles_scalable_2023,karras_analyzing_2024} offer a compelling path: a hospital trains a generative model locally and shares synthetic samples on demand, a privacy-preserving alternative to federated learning~\cite{rieke2020future,moroianu_improving_2025}. The practical value of this pipeline depends on whether generated images carry the discriminative features downstream classifiers need, particularly for tail classes. Current work addresses this through perceptual fidelity (FID~\cite{heusel2017gans}) and domain-specific autoencoder fine-tuning~\cite{varma_medvae_2025}, implicitly treating the latent space as a neutral intermediary. Recent studies on natural images challenge this assumption: latent spaces contain high-frequency artifacts~\cite{skorokhodov_improving_2025}, lack geometric equivariance~\cite{kouzelis_eq-vae_2025,zhou_alias-free_2025}, and benefit from alignment with pretrained representations~\cite{yu_representation_2025,leng_repa-e_2025,yao_reconstruction_2025,zheng_diffusion_2025,gui_adapting_2025}. The emerging view is that latent space structure actively shapes what a generative model can learn~\cite{bfl2025representation,falck_fourier_2025}, yet the implications for class-conditional generation in long-tailed medical settings remain unexplored.

We make this question precise through a systematic study across five autoencoder families and four clinical benchmarks. Reconstruction-space classifiers match image-space performance ($p{=}0.375$, Wilcoxon signed-rank test), confirming that the autoencoders faithfully preserve discriminative information. Yet latent-space classifiers suffer a substantial drop despite extensive architecture search, distillation, and noise-conditional training ($p = 9.5 \times 10^{-7}$, Wilcoxon across all 20 AE$\times$dataset pairs). We term this the \emph{learnability gap} (Fig.~\ref{fig:placeholder}): the information is present but structured in a way that resists direct learning. This gap propagates to generative augmentation: if the diffusion model cannot learn class-discriminative structure from the latents, its samples will lack the pathological markers classifiers rely on.

Our contributions are:
\textbf{(1)}~We identify and quantify the learnability gap across five autoencoder families and four medical benchmarks, showing it is systematic ($p = 9.5 \times 10^{-7}$) and independent of reconstruction fidelity, domain specificity, and initialization strategy.
\textbf{(2)}~We develop noise-conditioned latent classifiers with FiLM layers and image-space distillation that partially narrow the gap and offer $64{\times}$ throughput gains, making them practical for rejection sampling and quality filtering in generative pipelines.
\textbf{(3)}~We provide a controlled analysis isolating the sources of the gap through systematic ablations of fine-tuning, distillation, and noise conditioning, establishing latent space \emph{structure} as the bottleneck for long-tail medical synthesis.

\noindent\textbf{Related work.}
Long-tailed medical classification has been extensively benchmarked~\cite{holste_long-tailed_2022,holste_towards_2024,lin_cxr-lt_2025}, with LDAM~\cite{cao_learning_2019}, decoupled training~\cite{kang_decoupling_2020}, and focal loss~\cite{lin2017focal} as standard remedies. Generative augmentation via class-balancing diffusion~\cite{qin2023classbalancingdiffusionmodels}, guided adapters~\cite{nutzel2025grasp}, synthetic pretraining~\cite{moroianu_improving_2025}, and diversity control~\cite{dombrowski_image_2024} has shown promise. Latent diffusion~\cite{rombach2022high,ho2020denoising,peebles_scalable_2023,karras_analyzing_2024} relies on autoencoder latent spaces whose structure significantly impacts generation quality through high-frequency artifacts~\cite{skorokhodov_improving_2025}, equivariance deficits~\cite{kouzelis_eq-vae_2025,zhou_alias-free_2025}, and spectral biases~\cite{falck_fourier_2025}. Representation alignment methods~\cite{yu_representation_2025,leng_repa-e_2025,yao_reconstruction_2025,zheng_diffusion_2025,gui_adapting_2025} improve latent spaces using pretrained encoders such as DINOv2~\cite{oquab_dinov2_2024}, and latent-space learnability has been studied for natural images~\cite{bfl2025representation} and memorization detection~\cite{dombrowski2025lcmem}. Domain-specific autoencoders preserve clinical features~\cite{varma_medvae_2025}. Classifier guidance at inference~\cite{dhariwal2021diffusion,ho2022classifier} and autoguidance~\cite{karras_guiding_2024} improve sample quality but do not address the underlying representation. We differ by providing the first systematic diagnosis of the learnability gap in medical latent spaces and by developing noise-conditioned latent classifiers as both diagnostic and practical tools for generative pipelines.

%%------------

\section{Method}
\label{sec:method}

\begin{figure}[t]
    \centering
    \includegraphics[width=\linewidth]{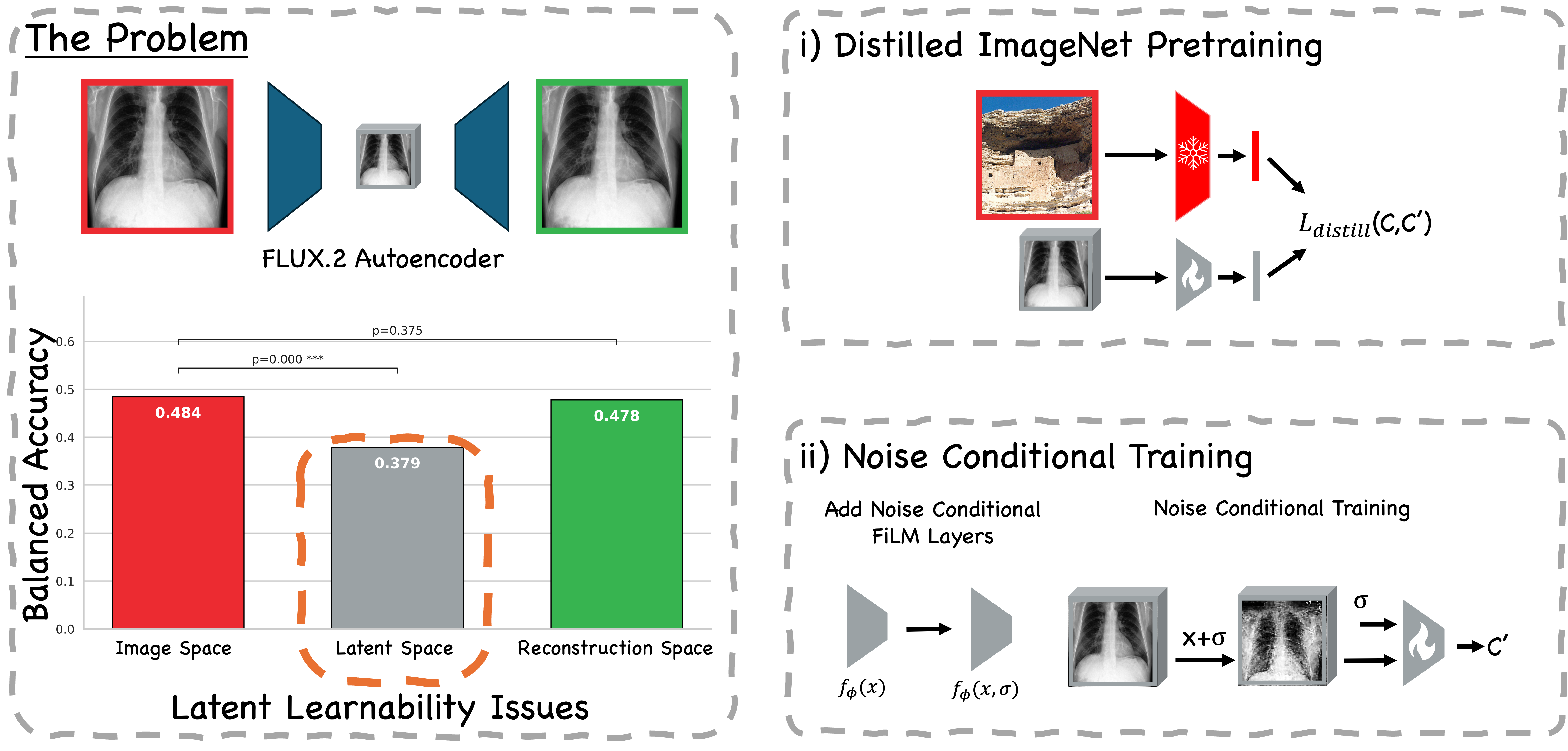}
    \caption{Method overview. A frozen pretrained autoencoder maps images to latent space and back. The learnability gap: reconstruction-space classifiers match image space, but latent-space classifiers degrade substantially.
A ConvNeXt-Tiny student classifier with FiLM-based noise conditioning is distilled from an image-space teacher. 
}
\label{fig:method}
    \label{fig:placeholder}
\end{figure}

An overview is given in Fig.~\ref{fig:method}. We operate in the latent space of a frozen pretrained autoencoder. Let $x_{\text{img}}$ denote an image, $x = E(x_{\text{img}})$ its latent encoding, and $\tilde{x}_{\text{img}} = G(x)$ the decoded reconstruction. All latents are channel-wise normalized over the training set.

\noindent\textbf{Three evaluation spaces.}
To isolate the effect of latent encoding from information loss, we define three evaluation spaces: \emph{image space} (classifiers trained on original images $x_{\text{img}}$), \emph{latent space} (classifiers trained on latent codes $x$), and \emph{reconstruction space} (classifiers trained on decoded reconstructions $\tilde{x}_{\text{img}} = G(E(x_{\text{img}}))$). If latent-space classifiers underperform reconstruction-space classifiers, the gap cannot be caused by missing information (since the decoder can recover it) but must instead reflect how the latent space \emph{structures} that information. This three-way comparison is the core diagnostic tool of our analysis.

\noindent\textbf{Noise-conditioned latent classifier.}
Our core component is a classifier $f_\phi(x_\sigma, \sigma)$ that predicts the class label $c$ from noisy latents at any noise level $\sigma$, forcing recovery of semantic information even at low signal-to-noise ratios. We adopt a ConvNeXt-Tiny~\cite{liu2022convnext} backbone operating directly on the latent tensor, chosen for its strong performance in prior latent-space studies~\cite{dombrowski2025lcmem}. The noise level $\sigma$ is embedded via Fourier features followed by an MLP with SiLU activations, and injected at the end of each ConvNeXt stage through Feature-wise Linear Modulation (FiLM)~\cite{perez2017filmvisualreasoninggeneral}: $\gamma, \beta = We(\sigma)$;\; $x' = x \odot (1+\gamma) + \beta$. FiLM parameters are zero-initialized so the backbone initially acts as an identity mapping and gradually learns noise-dependent modulation. The classifier is trained with $\mathcal{L}_{\text{cls}} = \mathbb{E}_{x,\varepsilon,\sigma}[\mathrm{CE}(f_\phi(x{+}\sigma\varepsilon,\,\sigma),\,c)]$.

Our motivation for noise conditioning is twofold. First, it improves classification by forcing the network to learn robust, noise-invariant features rather than overfitting to clean-latent statistics (Table~\ref{tab:ablation}). Second, noise-conditioned classifiers integrate naturally into diffusion model training as regularizers on the predicted clean latent $\hat{x}_0$ or as guidance models during sampling, providing a practical path toward narrowing the learnability gap at the generative  level.

\noindent\textbf{Distillation from image space.}
Direct training of the latent classifier is limited by the learnability gap itself. We therefore distill from a strong image-space teacher $T$ operating on reconstructions $\tilde{x}_{\text{img}}$. With teacher logits $z_T = T(\tilde{x}_{\text{img}})$ and student logits $z_S = f_\phi(x)$, we optimize:
\begin{equation}\label{eq:distill}
\mathcal{L}_{\text{distill}} = \alpha\,\mathrm{CE}(z_S, c) + (1-\alpha)\,\|z_S - z_T\|_2^2, \qquad \alpha = 0.5,
\end{equation}
combining ground-truth supervision with soft logit matching. This encourages the latent student to mimic the image-space decision boundary while retaining hard-label supervision. The combination of distillation and noise conditioning yields our strongest latent classifier, which we use both to quantify the learnability gap and as a practical component for downstream generative pipelines (\emph{e.g.}\ rejection sampling, training-time regularization).

\section{Experiments}
\label{sec:experiments}

\noindent\textbf{Setup.}
We evaluate on four medical benchmarks: MIMIC-CXR long-tail (19 chest X-ray findings, $N{=}111$k)~\cite{johnson_mimic-cxr_2019}, ISIC 2019 (8 skin lesion classes, $N{=}25$k)~\cite{codella2018skin,tschandl2018ham10000}, CT-RATE (13 CT findings, $N{=}23$k)~\cite{hamamci_developing_2025}, and Cardium (CHD vs.\ normal, $N{=}6.6$k)~\cite{vega2025cardiumcongenitalanomalyrecognition}. All datasets exhibit long-tailed distributions with imbalance ratios exceeding 50:1.
We compare five autoencoder families: Stable Diffusion 1.4 (SD~1.4), Flux.1~dev, Flux.2~dev, MedVAE~\cite{varma_medvae_2025}, and MedVAE fine-tuned on each target dataset (MedVAE-FT).
Image-space and reconstruction-space classifiers use ImageNet-pretrained ResNet-50~\cite{he2016deep} with LDAM loss~\cite{cao_learning_2019} and deferred reweighting at epoch 10, trained for up to 60 epochs (patience 15). Latent classifiers use ConvNeXt-Tiny~\cite{liu2022convnext} with the input layer adapted to each autoencoder's channel count. All results report 5-fold cross-validation on identical folds and seeds; we report mean$_{\pm\text{std}}$. We evaluate balanced accuracy (bACC), area under the ROC curve (AUC), and Matthews correlation coefficient (MCC).

%-------------------------------------------------------------
% TABLE 1: Main learnability gap
%-------------------------------------------------------------
\begin{table}[t]
\centering
\caption{The learnability gap across autoencoder families (pretrained init., 5-fold CV). Latent space (LS) classifiers underperform reconstruction space (RS), which matches image space. $^\dagger$\,Paired $t$-test $p{<}0.05$; $^\ddagger$\,$p{<}0.001$. Best LS per dataset in \textbf{bold}. Overall LS vs RS: Wilcoxon $p = 9.5 \times 10^{-7}$.}
\label{tab:learnability}
\resizebox{\textwidth}{!}{%
\setlength{\tabcolsep}{3.5pt}
\begin{tabular}{ll ccc ccc ccc ccc}
\toprule
& & \multicolumn{3}{c}{\textbf{CT-RATE}} & \multicolumn{3}{c}{\textbf{ISIC-2019}} & \multicolumn{3}{c}{\textbf{MIMIC-LT}} & \multicolumn{3}{c}{\textbf{Cardium}} \\
\cmidrule(lr){3-5}\cmidrule(lr){6-8}\cmidrule(lr){9-11}\cmidrule(lr){12-14}
Method & Sp. & bACC & AUC & MCC & bACC & AUC & MCC & bACC & AUC & MCC & bACC & AUC & MCC \\
\midrule
\rowcolor{lightgray}
Image Space & -- & $.255_{\pm.018}$ & $.720_{\pm.017}$ & $.280_{\pm.033}$ & $.749_{\pm.021}$ & $.947_{\pm.008}$ & $.703_{\pm.027}$ & $.263_{\pm.010}$ & $.760_{\pm.015}$ & $.279_{\pm.017}$ & $.669_{\pm.028}$ & $.743_{\pm.036}$ & $.280_{\pm.061}$ \\
\midrule
\multirow{2}{*}{SD 1.4}
 & LS & $.156_{\pm.010}^{\ddagger}$ & $.635_{\pm.020}$ & $.147_{\pm.014}$ & $.468_{\pm.012}^{\ddagger}$ & $.835_{\pm.011}$ & $.428_{\pm.020}$ & $.135_{\pm.007}^{\ddagger}$ & $.656_{\pm.013}$ & $.191_{\pm.022}$ & $.624_{\pm.014}$ & $.676_{\pm.010}$ & $.202_{\pm.012}$ \\
 & RS & $.246_{\pm.020}$ & $.727_{\pm.018}$ & $.257_{\pm.015}$ & $.746_{\pm.009}$ & $.943_{\pm.005}$ & $.700_{\pm.009}$ & $.261_{\pm.012}$ & $.769_{\pm.008}$ & $.285_{\pm.018}$ & $.662_{\pm.032}$ & $.723_{\pm.043}$ & $.270_{\pm.077}$ \\
\midrule
\multirow{2}{*}{Flux.1}
 & LS & $.167_{\pm.009}^{\ddagger}$ & $.639_{\pm.027}$ & $.153_{\pm.015}$ & $.496_{\pm.013}^{\ddagger}$ & $.853_{\pm.006}$ & $.466_{\pm.005}$ & $.132_{\pm.009}^{\ddagger}$ & $.662_{\pm.009}$ & $.197_{\pm.011}$ & $.627_{\pm.024}$ & $.682_{\pm.026}$ & $.209_{\pm.054}$ \\
 & RS & $.252_{\pm.020}$ & $.725_{\pm.014}$ & $.279_{\pm.017}$ & $.744_{\pm.017}$ & $.943_{\pm.004}$ & $.697_{\pm.015}$ & $.263_{\pm.025}$ & $.773_{\pm.008}$ & $.274_{\pm.023}$ & $.667_{\pm.036}$ & $.724_{\pm.055}$ & $.287_{\pm.068}$ \\
\midrule
\multirow{2}{*}{Flux.2}
 & LS & $\mathbf{.169}_{\pm.022}^{\dagger}$ & $.653_{\pm.019}$ & $\mathbf{.174}_{\pm.008}$ & $\mathbf{.552}_{\pm.013}^{\ddagger}$ & $.884_{\pm.003}$ & $\mathbf{.527}_{\pm.019}$ & $\mathbf{.166}_{\pm.010}^{\ddagger}$ & $.703_{\pm.005}$ & $\mathbf{.221}_{\pm.034}$ & $\mathbf{.639}_{\pm.026}$ & $.690_{\pm.053}$ & $\mathbf{.223}_{\pm.043}$ \\
 & RS & $.242_{\pm.017}$ & $.726_{\pm.011}$ & $.260_{\pm.008}$ & $.755_{\pm.019}$ & $.951_{\pm.005}$ & $.714_{\pm.012}$ & $.247_{\pm.023}$ & $.767_{\pm.008}$ & $.282_{\pm.014}$ & $.666_{\pm.021}$ & $.722_{\pm.036}$ & $.268_{\pm.038}$ \\
\midrule
\multirow{2}{*}{MedVAE}
 & LS & $.164_{\pm.014}^{\dagger}$ & $.641_{\pm.010}$ & $.176_{\pm.007}$ & $.539_{\pm.021}^{\ddagger}$ & $.877_{\pm.008}$ & $.509_{\pm.018}$ & $.144_{\pm.006}^{\ddagger}$ & $.673_{\pm.008}$ & $.210_{\pm.013}$ & $.633_{\pm.053}$ & $.684_{\pm.068}$ & $.227_{\pm.074}$ \\
 & RS & $.239_{\pm.015}$ & $.719_{\pm.011}$ & $.261_{\pm.017}$ & $.725_{\pm.018}$ & $.934_{\pm.007}$ & $.667_{\pm.013}$ & $.205_{\pm.011}$ & $.730_{\pm.011}$ & $.248_{\pm.029}$ & $.644_{\pm.028}$ & $.720_{\pm.051}$ & $.235_{\pm.059}$ \\
\midrule
\multirow{2}{*}{MedVAE-FT}
 & LS & $.164_{\pm.021}^{\dagger}$ & $.655_{\pm.010}$ & $.168_{\pm.018}$ & $.553_{\pm.015}^{\ddagger}$ & $.882_{\pm.006}$ & $.525_{\pm.013}$ & $.157_{\pm.006}^{\dagger}$ & $.682_{\pm.004}$ & $.205_{\pm.018}$ & $.623_{\pm.029}$ & $.680_{\pm.048}$ & $.203_{\pm.047}$ \\
 & RS & $.238_{\pm.019}$ & $.717_{\pm.019}$ & $.267_{\pm.011}$ & $.750_{\pm.021}$ & $.949_{\pm.008}$ & $.709_{\pm.018}$ & $.257_{\pm.027}$ & $.767_{\pm.013}$ & $.276_{\pm.017}$ & $.661_{\pm.046}$ & $.725_{\pm.048}$ & $.265_{\pm.074}$ \\
\bottomrule
\end{tabular}}
\end{table}

\noindent\textbf{The learnability gap is large and consistent.}
Table~\ref{tab:learnability} presents the central finding. Across all five autoencoder families and four datasets, latent-space classifiers underperform their reconstruction-space counterparts. On ISIC-2019, the bACC gap exceeds 18 percentage points for every autoencoder (up to 27.8 for SD~1.4; $p{<}10^{-4}$, paired $t$-test). On MIMIC-LT the gap ranges from 6 to 13 points, consistently significant ($p{<}0.002$). A Wilcoxon signed-rank test across all 20 AE$\times$dataset pairs confirms the gap is systematic ($p = 9.5 \times 10^{-7}$). Reconstruction-space performance is not significantly different from image space ($p{=}0.375$, Wilcoxon), confirming that discriminative information is faithfully preserved. Cardium is the only dataset where the gap does not reach significance, attributable to its binary nature and high cross-fold variance.

%-------------------------------------------------------------
% TABLES 2 + 6 SIDE BY SIDE
%-------------------------------------------------------------
\begin{table}[t]
\centering
\begin{minipage}[t]{0.56\textwidth}
\centering
\caption{Recon. quality. High fidelity does not predict latent-space learnability: Flux.2 leads in PSNR but shows the largest LS gap on ISIC.}
\label{tab:recon}
\setlength{\tabcolsep}{2.5pt}
\resizebox{\textwidth}{!}{%
\begin{tabular}{l cc cc cc cc}
\toprule
& \multicolumn{2}{c}{CT-RATE} & \multicolumn{2}{c}{ISIC} & \multicolumn{2}{c}{MIMIC} & \multicolumn{2}{c}{Cardium} \\
\cmidrule(lr){2-3}\cmidrule(lr){4-5}\cmidrule(lr){6-7}\cmidrule(lr){8-9}
& SSIM & PSNR & SSIM & PSNR & SSIM & PSNR & SSIM & PSNR \\
\midrule
SD 1.4     & .924 & 34.2 & .971 & 37.3 & .983 & 35.2 & .985 & 30.5 \\
Flux.1     & .946 & 40.8 & .994 & 40.3 & .997 & 40.5 & .999 & 39.1 \\
Flux.2     & .941 & 40.6 & \textbf{.994} & \textbf{42.3} & \textbf{.997} & \textbf{41.9} & \textbf{.999} & \textbf{40.8} \\
MedVAE     & .938 & 38.2 & .963 & 24.9 & .711 & 23.7 & .991 & 32.7 \\
MedVAE-FT  & .939 & 39.5 & .991 & 41.1 & .996 & 39.8 & .996 & 35.2 \\
\bottomrule
\end{tabular}}
\end{minipage}%
\hfill
\begin{minipage}[t]{0.43\textwidth}
\centering
\caption{ImageNet, 24\,h budget, single GH200. Latent training: $64{\times}$ faster, $120{\times}$ less memory.}
\label{tab:imagenet}
\setlength{\tabcolsep}{2.5pt}
\resizebox{\textwidth}{!}{%
\begin{tabular}{lccccc}
\toprule
Space & bACC & AUC & MCC & S/s & MB \\
\midrule
Image        & .459 & --   & --   & 50   & 20k \\
Latent       & .630 & .996 & .629 & 3.2k & 166 \\
\;+ Distill. & .711 & .993 & .710 & 3.2k & 166 \\
\bottomrule
\end{tabular}}%
\end{minipage}
\end{table}

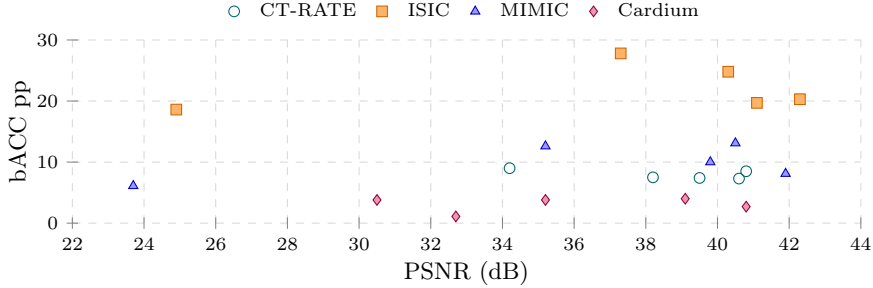
\begin{figure}[t]
    \centering
    \begin{tikzpicture}
    \begin{axis}[
        width=12cm, height=4cm, % Sleek, wide, and short aspect ratio (approx 4cm high)
        xlabel={PSNR (dB)},
        ylabel={bACC pp},
        xlabel style={font=\footnotesize, yshift=2pt},
        ylabel style={font=\footnotesize, yshift=-4pt},
        xmin=22, xmax=44,
        ymin=0, ymax=30,
        x tick label style={font=\scriptsize},
        y tick label style={font=\scriptsize},
        % Modern, clean MICCAI/publication style
        axis line style={draw=none},
        axis x line=bottom,
        axis y line=left,
        grid=major,
        grid style={dashed, gray!30},
        % Horizontal legend moved outside to save vertical space
        legend style={
            at={(0.5, 1.05)},
            anchor=south,
            font=\scriptsize,
            draw=none,
            fill=none,
            legend columns=-1, % Forces single row
            column sep=0.2cm
        },
        scatter/classes={
            ct={mark=o, draw=teal!80!black, fill=teal!50},
            isic={mark=square*, draw=orange!80!black, fill=orange!60},
            mimic={mark=triangle*, draw=blue!80!black, fill=blue!40},
            card={mark=diamond*, draw=purple!80!black, fill=purple!40}
        },
    ]
    % CT-RATE points
    \addplot[scatter, only marks, mark size=2pt, scatter src=explicit symbolic]
        coordinates {
            (34.2, 9.0) [ct]
            (40.8, 8.5) [ct]
            (40.6, 7.3) [ct]
            (38.2, 7.5) [ct]
            (39.5, 7.4) [ct]
        };
    % ISIC points
    \addplot[scatter, only marks, mark size=2pt, scatter src=explicit symbolic]
        coordinates {
            (37.3, 27.8) [isic]
            (40.3, 24.8) [isic]
            (42.3, 20.3) [isic]
            (24.9, 18.6) [isic]
            (41.1, 19.7) [isic]
        };
    % MIMIC points
    \addplot[scatter, only marks, mark size=2pt, scatter src=explicit symbolic]
        coordinates {
            (35.2, 12.6) [mimic]
            (40.5, 13.1) [mimic]
            (41.9, 8.1) [mimic]
            (23.7, 6.1) [mimic]
            (39.8, 10.0) [mimic]
        };
    % Cardium points
    \addplot[scatter, only marks, mark size=2pt, scatter src=explicit symbolic]
        coordinates {
            (30.5, 3.8) [card]
            (39.1, 4.0) [card]
            (40.8, 2.7) [card]
            (32.7, 1.1) [card]
            (35.2, 3.8) [card]
        };
    \legend{CT-RATE, ISIC, MIMIC, Cardium}
    \end{axis}
    \end{tikzpicture}
    \vspace{-5pt}
    \caption{PSNR vs.\ learnability gap. Each point is one AE$\times$dataset pair. Higher fidelity does not reduce the gap; dataset difficulty dominates.}
    \label{fig:scatter}
\end{figure}

\begin{figure}[t]
    \centering
    \includegraphics[width=\linewidth]{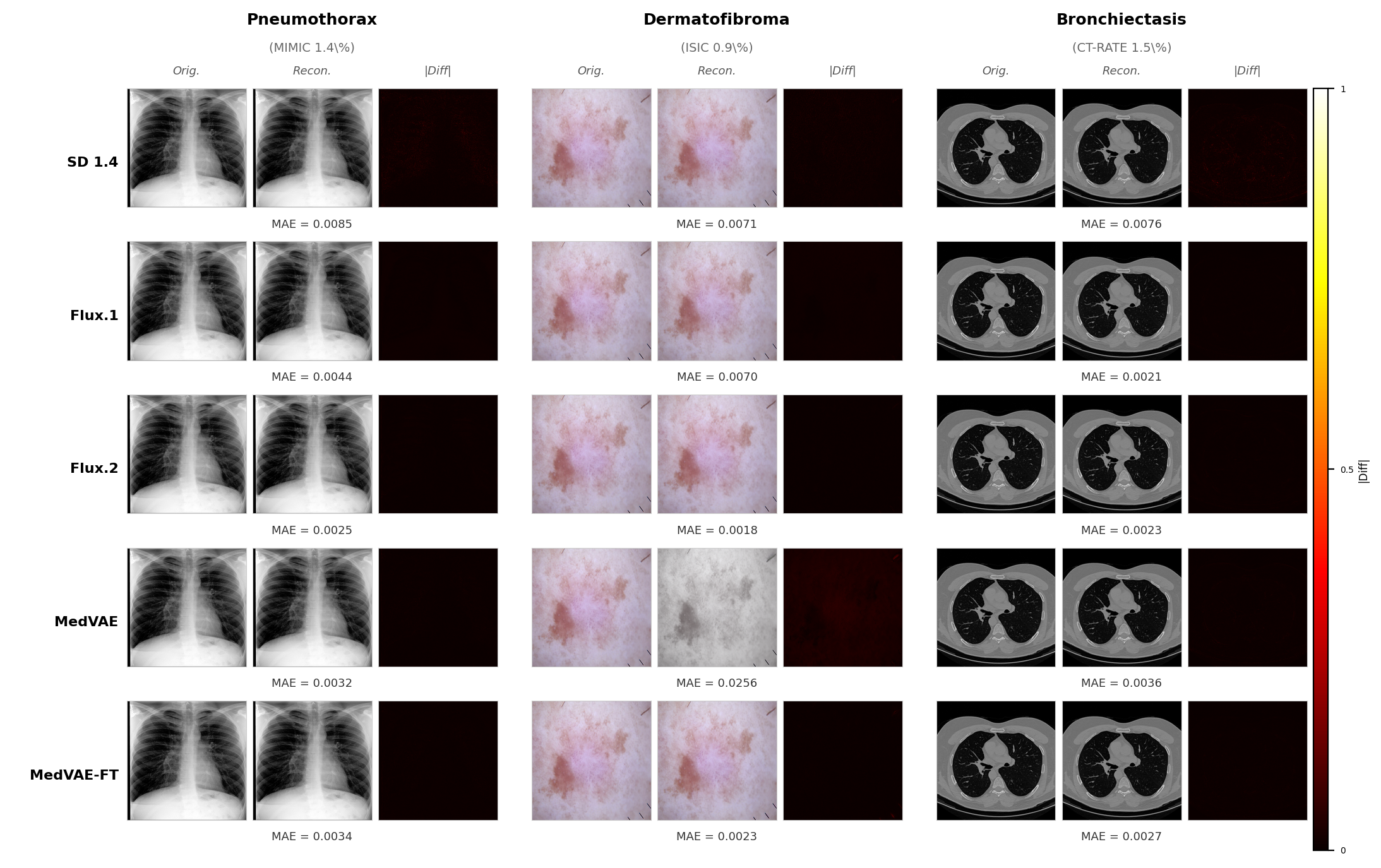}
    \caption{Reconstruction samples and pixel-wise absolute difference for long-tail classes across CT-RATE, MIMIC-LT, and ISIC. Reconstructions are near-identical to originals, confirming that the autoencoder preserves discriminative information despite the large classification gap in latent space.}
    \label{fig:qualitativesamples}
\end{figure}

\noindent\textbf{Reconstruction fidelity does not predict learnability.}
Table~\ref{tab:recon} reports reconstruction quality for each autoencoder. Flux.2 achieves the highest PSNR (41.4\,dB macro average) yet also exhibits a large LS gap, while MedVAE sometimes has lower fidelity (\emph{e.g.}\ 23.7\,dB on MIMIC) but a comparable classification gap. Fig.~\ref{fig:qualitativesamples} confirms this visually: pixel-wise differences between originals and reconstructions are minimal across autoencoders, yet the classification gap persists. Fig.~\ref{fig:scatter} makes this explicit: across all 20 AE$\times$dataset combinations, PSNR shows no predictive relationship with gap magnitude. This rules out information loss as the cause and implicates the \emph{structure} of the latent space.

\noindent\textbf{Medical fine-tuning does not close the gap.}
Comparing MedVAE with MedVAE-FT in Table~\ref{tab:learnability} reveals that fine-tuning the autoencoder on in-domain data improves reconstruction quality substantially (PSNR from 29.9 to 38.9\,dB on average, Table~\ref{tab:recon}) but does not meaningfully improve latent-space classification: the macro-average LS bACC changes from .370 to .374, with significance on only one of four datasets (MIMIC-LT, $p{=}0.04$; paired $t$-test). Meanwhile, RS benefits more from fine-tuning (MIMIC $p{=}0.008$, ISIC $p{=}0.02$), widening the gap rather than closing it. This confirms that the learnability gap is structural, not a domain mismatch, and that investing compute in autoencoder fine-tuning is not an effective remedy.

%-------------------------------------------------------------
% TABLE: Expanded ablation with AUC and MCC
%-------------------------------------------------------------
\begin{table}[t]
\centering
\caption{Latent classifier ablation on Flux.2 (5-fold CV). Each component helps incrementally, but the gap to RS persists, confirming its structural nature. $^\star$\,Significant vs.\ na\"ive ($p{<}0.05$, paired $t$-test).}
\label{tab:ablation}
\setlength{\tabcolsep}{2.5pt}
\small
\resizebox{\textwidth}{!}{%
\begin{tabular}{l ccc ccc ccc ccc c}
\toprule
& \multicolumn{3}{c}{CT-RATE} & \multicolumn{3}{c}{ISIC-2019} & \multicolumn{3}{c}{MIMIC-LT} & \multicolumn{3}{c}{Cardium} & Macro \\
\cmidrule(lr){2-4}\cmidrule(lr){5-7}\cmidrule(lr){8-10}\cmidrule(lr){11-13}\cmidrule(lr){14-14}
Method & bACC & AUC & MCC & bACC & AUC & MCC & bACC & AUC & MCC & bACC & AUC & MCC & AUC \\
\midrule
\rowcolor{lightgray}
RS baseline & $.242_{\pm.017}$ & $.726_{\pm.011}$ & $.260_{\pm.008}$ & $.755_{\pm.019}$ & $.951_{\pm.005}$ & $.714_{\pm.012}$ & $.247_{\pm.023}$ & $.767_{\pm.008}$ & $.282_{\pm.014}$ & $.666_{\pm.021}$ & $.722_{\pm.036}$ & $.268_{\pm.038}$ & .791 \\
\midrule
Na\"ive LS & $.169_{\pm.022}$ & $.653_{\pm.019}$ & $.174_{\pm.008}$ & $.552_{\pm.013}$ & $.884_{\pm.003}$ & $.527_{\pm.019}$ & $.166_{\pm.010}$ & $.703_{\pm.005}$ & $.221_{\pm.034}$ & $.639_{\pm.026}$ & $.690_{\pm.053}$ & $.223_{\pm.043}$ & .731 \\
+ Distill. & $.174_{\pm.007}$ & $.664_{\pm.018}$ & $.202_{\pm.015}$ & $.581_{\pm.064}$ & $.898_{\pm.021}$ & $.565_{\pm.045}$ & $.152_{\pm.005}$ & $.697_{\pm.017}$ & $.228_{\pm.029}$ & $.653_{\pm.024}$ & $.728_{\pm.028}$ & $.260_{\pm.040}$ & .747 \\
+ HP opt.$^\star$ & $.205_{\pm.015}$ & $.695_{\pm.018}$ & $.209_{\pm.018}$ & $.643_{\pm.021}$ & $.901_{\pm.012}$ & $.584_{\pm.039}$ & $.202_{\pm.010}$ & $.740_{\pm.007}$ & $.265_{\pm.022}$ & $.567_{\pm.093}$ & $.573_{\pm.159}$ & $.109_{\pm.151}$ & .727 \\
+ Noise cond. & $.174_{\pm.028}$ & $.669_{\pm.027}$ & $.203_{\pm.028}$ & $.591_{\pm.043}$ & $.896_{\pm.008}$ & $.563_{\pm.020}$ & $.177_{\pm.011}$ & $.720_{\pm.014}$ & $.258_{\pm.012}$ & $.663_{\pm.020}$ & $.733_{\pm.009}$ & $.289_{\pm.021}$ & .754 \\
\bottomrule
\end{tabular}}
\end{table}

\noindent\textbf{Latent classifier ablation.} 
Table~\ref{tab:ablation} ablates strategies for training latent-space classifiers on Flux.2. Na\"ive training already exposes the full learnability gap. Systematic hyperparameter optimization yields a distilled variant reaching a macro bACC of .404 ($p{<}0.03$ on 3 of 4 datasets), though it remains unstable on the data-limited Cardium task ($bACC=.567_{\pm.093}$). 
Introducing noise conditioning via FiLM layers slightly reduces macro bACC to .401 but substantially improves discriminative robustness: mean AUC rises from .727 to .754 and MCC from .292 to .328. This improvement is most notable on Cardium (MCC $.223\to.289$), confirming that noise-aware training learns more robust decision boundaries. Despite these interventions, a gap of 4--11 bACC points relative to reconstruction space (RS) persists. This suggests the learnability gap is a structural property of the latent space rather than an artifact of suboptimal training.

\noindent\textbf{Latent classifiers are fast.}
Table~\ref{tab:imagenet} demonstrates that latent classifiers, despite the learnability gap, offer compelling practical advantages. On ImageNet with a fixed 24-hour budget on a single NVIDIA GH200, latent ConvNeXt-Tiny reaches bACC .630 vs.\ .459 in image space, a 17-point gain from operating in the compressed domain. With distillation it reaches .711. Throughput increases $64{\times}$ (3.2k vs.\ 50 samples/sec) and memory drops $120{\times}$ (166\,MB vs.\ 20\,GB) compared to image space classifiers which have to decode the latents first. This makes latent classifiers practical as fast quality filters for generated images and as components in training-time regularization schemes for diffusion models.

\noindent\textbf{Why does the gap exist?}
Our experiments systematically rule out several potential explanations. Information loss is excluded by the RS$\approx$IS equivalence ($p{=}0.375$). Domain mismatch is excluded by the failure of medical fine-tuning to close the gap despite improving reconstruction from 29.9 to 38.9\,dB PSNR. Insufficient classifier capacity is unlikely: ConvNeXt-Tiny achieves strong RS and image-space performance, and the gap persists or widens with stronger classifiers and extensive hyperparameter tuning. We hypothesize that autoencoders trained for pixel-wise reconstruction distribute class-discriminative features across high-frequency spatial patterns and inter-channel correlations that convolutional classifiers struggle to exploit, while the decoder inverts these patterns precisely because it was jointly trained with the encoder. This interpretation is consistent with recent findings on spectral biases~\cite{falck_fourier_2025} and high-frequency artifacts~\cite{skorokhodov_improving_2025} in autoencoder latent spaces. The gap is largest on ISIC-2019 (18-28 points), where fine-grained texture differences between lesion types may be particularly susceptible to such encoding artifacts. These findings suggest that latent space restructuring, \emph{e.g.}\ through class-conditional objectives, equivariance-aware architectures~\cite{kouzelis_eq-vae_2025}, or alignment with discriminative encoders~\cite{yu_representation_2025}, which is a more promising direction than improving reconstruction fidelity.

% ---------------------------------------------------------------
% DISCUSSION
% ---------------------------------------------------------------
\noindent\textbf{Discussion.}
Our results establish that the bottleneck in latent diffusion for medical long-tail synthesis is not the autoencoder's representational capacity but the \emph{learnability} of its latent space. The gap is robust across five autoencoder families, four clinical modalities, and extensive hyperparameter tuning ($p = 9.5 \times 10^{-7}$, Wilcoxon). This finding has direct implications for the field: efforts to improve synthetic medical data should prioritize latent space restructuring over reconstruction fidelity or domain-specific fine-tuning. Our noise-conditioned latent classifiers, while not fully closing the gap, provide both a diagnostic framework for evaluating latent space quality and practical components for generative pipelines through their $64{\times}$ throughput advantage. A limitation is that we diagnose the gap but do not resolve it at the representation level. Removing this gap and validating with expert radiologist evaluation are natural next steps.

\noindent\textbf{Broader impact.}
Our findings have practical implications for the deployment of generative models in clinical settings. The learnability gap means that current latent diffusion pipelines may systematically underrepresent rare pathologies in synthetic training data, potentially perpetuating the very diagnostic biases they are designed to mitigate. By identifying latent space structure as the bottleneck, we redirect optimization efforts away from expensive autoencoder fine-tuning toward more targeted interventions. Our noise-conditioned latent classifiers can serve as lightweight quality filters in privacy-preserving synthetic data sharing pipelines, enabling hospitals to verify the clinical plausibility of generated images without exposing real patient data. Addressing the learnability gap could accelerate the adoption of generative augmentation for rare disease diagnosis, where data scarcity is most acute and the clinical need is greatest.

\section{Conclusion}

We introduced the \emph{learnability gap}: a systematic discrepancy between the discriminative information preserved in autoencoder latent spaces and the ability of downstream models to access it. Through a comprehensive exploration spanning five autoencoder families, four long-tailed medical benchmarks, and multiple classifier training paradigms, we showed that this gap, \emph{i.e.} not reconstruction fidelity or domain specificity, is the primary bottleneck limiting the utility of latent representations for medical data augmentation. The gap is highly significant ($p = 9.5 \times 10^{-7}$) and is not remedied by medical-domain autoencoder fine-tuning. We developed noise-conditioned latent classifiers with FiLM layers and image-space distillation that partially narrow the gap and offer dramatic efficiency gains ($64{\times}$ throughput, $120{\times}$ memory reduction), making them practical for quality filtering and rejection sampling in generative pipelines. We provide three actionable insights: (i)~large-scale pretrained autoencoders are sufficient for medical synthesis without domain fine-tuning, as the bottleneck lies in latent space structure rather than domain specificity; (ii)~latent-space classifiers are fast and useful but require careful training via distillation and noise conditioning; and (iii)~the persistent learnability gap identifies latent space restructuring as the critical open problem for advancing synthetic medical data augmentation.

\begin{credits}
\subsubsection{\ackname} We acknowledge HPC resources from NHR@FAU (projects b143dc, b180dc), funded by federal and Bavarian state authorities and Gerhard Wellein's HPC approach. NHR@FAU hardware is partially funded by DFG 440719683. Additional support was received from ERC projects MIA-NORMAL 101083647, DFG 513220538 and 512819079, and the state of Bavaria (HTA and the Bavarian Foundation Model Initiative). We further acknowledge resources provided by the Isambard-AI National AI Research Resource (AIRR), operated by the University of Bristol and funded by DSIT via UKRI and STFC [ST/AIRR/I-A-I/1023]~\cite{mcintoshsmith2024isambardai}. We were supported by coding agents and LLMs from Anthropic, OpenAI, Google, and Mistral AI, for text polishing, coding, experiment orchestration, and cluster monitoring.

\subsubsection{\discintname}
The authors have no relevant competing interests.
\end{credits}

%\newpage
\bibliographystyle{splncs04}
\bibliography{main}

\end{document}